\documentclass{article}

\usepackage{ArXiv_style}
\usepackage[british]{babel}
\usepackage[utf8]{inputenc} 
\usepackage[T1]{fontenc}    
\usepackage{url}            
\usepackage{booktabs}       
\usepackage{amsfonts}       
\usepackage{nicefrac}       
\usepackage{microtype}      
\usepackage{xcolor}         

\usepackage{natbib} 
    \renewcommand{\bibsection}{\subsubsection*{References}}

\usepackage[%
linewidth=1pt,
middlelinecolor= black,
middlelinewidth=0.4pt,
roundcorner=1pt,
topline = false,
rightline = false,
bottomline = false,
rightmargin=0pt,
skipbelow=0pt,
leftmargin=0cm,
innerleftmargin=0.25cm,
innerrightmargin=0pt,
innertopmargin=0pt,
innerbottommargin=0pt,
]{mdframed}
\usepackage[utf8]{inputenc} 
\usepackage[T1]{fontenc}    
\usepackage[hidelinks]{hyperref}       
\usepackage{url}            
\usepackage{booktabs}       
\usepackage{amsfonts}       
\usepackage{nicefrac}       
\usepackage{microtype}      
\usepackage{float}
\usepackage{courier}
\usepackage{listings, lstautogobble,amsfonts}
\usepackage{amsmath,amssymb,amsthm}
\usepackage{mathtools}
\usepackage{caption}
\usepackage{subcaption}
\usepackage{xspace}
\usepackage{xcolor}
\usepackage{graphicx}
\usepackage[normalem]{ulem}
\usepackage{caption}
\usepackage{multicol}
\usepackage{bbm}
\usepackage{thmtools}
\usepackage{bm}
\usepackage{thm-restate}
 \usepackage{todonotes}
\usepackage[inline]{enumitem}
\usepackage{soul}
\usepackage{inconsolata}
\usepackage{physics}
\usepackage{wrapfig}
\usepackage{stmaryrd}
\usepackage{relsize}

\usepackage{dsfont}
\usepackage{multirow}

\usepackage{pgfplots}
\pgfplotsset{compat=1.18} 

\usepackage{tikz}
\usepackage{circuitikz}
\usetikzlibrary{
    calc, 
    shapes,
    shapes.misc,
    shapes.symbols,
    backgrounds,
    fit,
    arrows,
    arrows.meta,
    positioning,
    decorations.pathmorphing,
    patterns,
    patterns.meta,
    scopes,
}

\usepackage[%
    linewidth=1pt,
    middlelinecolor= black,
    middlelinewidth=0.4pt,
    roundcorner=1pt,
    topline = false,
    rightline = false,
    bottomline = false,
    rightmargin=0pt,
    skipbelow=0pt,
    leftmargin=0cm,
    innerleftmargin=0.25cm,
    innerrightmargin=0pt,
    innertopmargin=0pt,
    innerbottommargin=0pt,
]{mdframed}

\usepackage[defaultsans]{lato}

\theoremstyle{definition}
\newtheorem{theorem}{Theorem}[section]
\theoremstyle{definition}
\newtheorem{definition}[theorem]{Definition}
\theoremstyle{definition}

\theoremstyle{definition}
\newtheorem{claim}[theorem]{Claim}
\theoremstyle{definition}
\newtheorem{proposition}[theorem]{Proposition}
\theoremstyle{definition}
\newtheorem{example}[theorem]{Example}
\theoremstyle{definition}

\theoremstyle{definition}

\newcommand{\operators}{\ensuremath{\mathcal{O}}\xspace}

\newcommand{\Exp}[2]{\ensuremath{\mathbb{E}_{#1}\left[#2\right]}}

\newcommand{\indicator}{\ensuremath{\mathbbm{1}}}
\newcommand{\ive}[1]{\llbracket#1\rrbracket}

\definecolor{red_salsa}{HTML}{F94144}
\definecolor{orange_red}{HTML}{F3722C}
\definecolor{yellow_orange}{HTML}{F8961E}
\definecolor{mango_tango}{HTML}{F9844A}
\definecolor{maize_crayola}{HTML}{F9C74F}
\definecolor{pistachio}{HTML}{90BE6D}
\definecolor{jungle_green}{HTML}{43AA8B}
\definecolor{steel_teal}{HTML}{4D908E}
\definecolor{queen_blue}{HTML}{577590}
\definecolor{celadon_blue}{HTML}{277DA1}
\definecolor{softer_black}{HTML}{2A2A2A}

\definecolor{poinant_purple}{HTML}{A87DC1}
\definecolor{purpose_purple}{HTML}{4D5AC0}
\definecolor{bliss_brown}{HTML}{C36D3A}
\definecolor{turd_turqoise}{HTML}{44BAC3}
\definecolor{soft_white}{HTML}{F1F1F1}
\definecolor{softer_white}{HTML}{E5E5E5}
\definecolor{grey}{HTML}{A5A5A5}
\definecolor{soft_black}{HTML}{101010}
\definecolor{softer_black}{HTML}{2A2A2A}

\newcommand{\symbols}{\ensuremath{\mathcal{S}}\xspace}

\newcommand{\atoms}{\ensuremath{\mathcal{A}}\xspace}
\newcommand{\semvalues}{\ensuremath{V}\xspace}

\newcommand{\sigmaalg}{$\sigma$-algebra\xspace}

\newcommand{\params}{\ensuremath{\boldsymbol{\theta}}}
\newcommand{\q}[1]{``#1''}

\newcommand{\colouraccent}[2]{\textcolor{#1}{\textbf{#2}}}

\renewcommand{\qq}[1]{``#1''}
\newcommand{\qemph}[1]{\emph{``#1''}}

\newcommand{\colournode}[4]{%
    \tikz[baseline=(#3.base), remember picture]
    \node[
        rounded corners=0.25em, 
        fill=#1, 
        fill opacity=0.2,
        text opacity=1,
        inner sep=#4,
        outer sep=0em
    ] (#3) {
        #2%
    };
    \xspace
}

\newsavebox{\imagebox}

\newenvironment{argument}{\raggedright\emph{Argument. }}{\hfill$\square$}

\tikzset{
    circnode/.style={
        fill=grey, align=center,
        fill opacity=0.25, text opacity=1,
        circle, 
        minimum width=2em
    },
    sqnode/.style={
        fill=grey,
        fill opacity=0.25, text opacity=1,
        rounded corners=0.25em,align=center,
        minimum width=2.5em, minimum height=2.5em
    },
    regpath/.style={
        -Triangle[round], draw=softer_black, ultra thick, rounded corners=0.25em
    },
    thinpath/.style={
        -Triangle[round], draw=softer_black, thick, rounded corners=0.25em
    },
    thinline/.style={
        solid, draw=softer_black, thick, rounded corners=0.25em, shorten <= 1mm, shorten >= 1mm
    },
    between/.style args={#1 and #2}{
         at = ($(#1)!0.5!(#2)$)
    }
}

\title{Defining neurosymbolic AI}

\author{
\href{mailto:lennert.desmet@kuleuven.be}{Lennert De Smet} \\
Department of Computer Science, \\
KU Leuven, Belgium \\
\url{lennert.desmet@kuleuven.be}
\And
\href{mailto:luc.deraedt@kuleuven.be}{Luc De Raedt} \\
Department of Computer Science, \\
KU Leuven, Belgium \\
and \"Orebro University, Sweden \\
\selectfont\url{luc.deraedt@kuleuven.be}
}

\numberwithin{equation}{section}

\begin{document}

\maketitle
\thispagestyle{fancy}

\begin{abstract}
    Neurosymbolic AI focuses on integrating learning and reasoning, in particular, on unifying logical  and neural representations. Despite the existence of an alphabet soup of neurosymbolic AI systems, the field is lacking a generally accepted formal definition of what neurosymbolic models and inference really are.  We introduce a formal definition for neurosymbolic AI that makes abstraction of its key ingredients. More specifically, we define neurosymbolic inference as the computation of an integral over a product of a logical and a belief function. We show that our neurosymbolic AI definition makes abstraction of  key representative neurosymbolic AI systems.
\end{abstract}

\section{Introduction}

Neurosymbolic AI (NeSy) combines more traditional symbolic AI techniques with the latest advances in deep learning.
It integrates neural and numerical data processing with symbolic reasoning and background knowledge. 
The advantages of this combination have already been demonstrated in applications such as providing safety guarantees~\citep{yang2023safe}, learning from distant signals and supervision by deduction~\citep{augustine2022visual}, and more~\citep{delong2023neurosymbolic,jiao2024valid,sun2024neurosymbolic}.
Neurosymbolic AI  is attracting a lot of attention~\citep{hochreiter,besold2017neural,garcez23,de2020statistical,box,garcez2019neural,bader2005dimensions,van_bekkum_modular_2021,dash2021tell} and has been termed  \qemph{the most promising approach to a broad AI} by Hochreiter~\citep{hochreiter} and the \qemph{3rd wave in AI} by Garcez and Lamb~\citep{garcez23}. 
It is mentioned as an innovation trigger on Gartner's hype cycle\footnote{\url{https://www.gartner.com/en/articles/hype-cycle-for-artificial-intelligence}} 
and there are now dedicated journals\footnote{\url{https://neurosymbolic-ai-journal.com/}}, conferences\footnote{\url{https://2025.nesyconf.org/}}, and summer schools\footnote{\url{https://neurosymbolic.github.io/nsss2024/}} devoted to neurosymbolic AI.  
  
Despite the wide interest, the term neurosymbolic AI is used for many different types of integrations
of neural and symbolic AI systems.
For instance, Henry Kautz describes six different types of such integrations~\citep{Kautz}, some requiring tighter interfaces between the neural and the symbolic component, others looser ones.
As Garcez and Lamb~\citep{garcez23}, we will focus on the original and stricter interpretation of the term neurosymbolic AI, which Garcez and Lamb describe as \qemph{research that integrates in a principled way neural network-based learning with symbolic knowledge representation and logical reasoning}. 
This is also the dominant view in the Neurosymbolic AI Journal and  Conference. 
Within this view there exist numerous models, systems and techniques that integrate logic with neural network-based approaches, see for instance \citep{marra2024statistical, garcez2022neural,besold2017neural,Hitzler}  for overviews. 
However, the focus in the field is very much on designing bespoke systems that score best on the latest benchmarks, which results in an alphabet soup of systems.
This comes at the expense of understanding the underlying principles and commonalities that these systems share, which is hindering progress in the field.
What is lacking is a commonly agreed formal definition and framework for specifying,  comparing  and developing neurosymbolic AI models and problems.
It is precisely this gap that this paper wants to bridge.  

More specifically, we contribute formal definitions 
of a neurosymbolic model and neurosymbolic inference. 
These definitions are based on the observation that the vast majority of neurosymbolic AI models combine {\em logic} with {\em beliefs}.
The term logic refers to the wide variety of logics that are used in neurosymbolic AI ranging from Boolean logic to first-order and fuzzy logic, as well as their combinations.
The term belief refers to a weighting component used by many neurosymbolic AI models~\citep{manhaeve_neural_2021,winters2021deepstochlog,pryor2022neupsl,yang2020neurasp} often derived from
statistical relational AI models \citep{DeRaedtKerstingEtAl16}. 
Within our framework, neurosymbolic inference can be formally defined as aggregation (or marginalisation) of a product of a logic and a belief function.
Our definitions provide a semantic framework for NeSy models that clarifies their components and the way they interact.
We will show that many representative classes of NeSy models and tasks, including those based on probabilistic logic \citep{manhaeve_neural_2021,yang2020neurasp}, fuzzy logic \citep{LTN}, or soft logic \citep{pryor2022neupsl},
can be cast within our definitions by instantiating the logic, the belief function and the neural networks appropriately.
As a consequence, our definitions can be used to relate, compare and develop different NeSy models in a principled manner, as well as to study fundamental properties of NeSy models and tasks.

\section{Logic as the symbol level}

In  Garcez and Lamb's perspective on neurosymbolic AI, the symbol level is viewed as symbolic knowledge representation and logical reasoning. 
We will therefore focus on using logical languages,  although the proposed definitions in principle apply to other formal languages and automata~\citep{de2021compositional, manginas2024nesya}.
We will also allow for a wide range of semantics for these languages, in order to support Boolean, fuzzy and other logics. 

More formally, a language $L$ is a set of sentences over an ordered set $\symbols = \left\{s_i\right\}_{i \in I}$ of symbols $s_i$ with domains $D_i \subseteq \mathbb{D}$ that can be embedded in a shared domain $\mathbb{D}$ and interact with operators \operators.~\footnote{The assumption of having a shared domain is made to simplify notation and holds without loss of generality.}
Symbols are assigned values by an interpretation $\omega: \symbols \to \mathbb{D}$ and we use $\Omega = \mathbb{D}^{\symbols}$ to denote the set of all possible interpretations over the symbols $\symbols$.
Given an interpretation $\omega \in \Omega$ and a sentence $\varphi \in L$, the \emph{semantics} $\mu: L \times \Omega \rightarrow \semvalues$ of the language $L$ maps $\varphi$ to a semantic value $\mu(\varphi, \omega)$ in the interpretation $\omega$, which we will often denote as $\varphi(\omega)$  for brevity.
The set \semvalues of semantic values is assumed to be embeddable in $\mathbb{R}^+$.~\footnote{This assumption also simplifies notation and is not a hard one, e.g. it is common to represent Boolean truth values $\top$ and falsehood $\bot$ as 1 and 0.}
If a symbol $s_i$ has a domain $D_i$ that is equal to the set \semvalues of semantic values, then it will be called an \emph{atom symbol} and the set of all atom symbols will be denoted as \atoms.
Atom symbols are special as they are directly assigned a meaningful value in an interpretation.

We illustrate the  different choices of language and semantics using both fuzzy~\citep{zadeh1965fuzzy,ruspini1991semantics} and Boolean semantics for propositional logic and SMT logic sentences.

\begin{example}[Boolean propositional logic]\label{example:boolean_propositional}
Propositional logic is the language that consists of sentences over symbols \symbols called \emph{propositions} that can be connected with logical operators such as $\Rightarrow$ and $\lor$.
For example, the sentence $\varphi$ equivalent to
\begin{align}
    \label{eq:main_example}
    \texttt{happy} \Rightarrow \texttt{(coffee} \lor \texttt{publication)},
\end{align}
states that happiness \texttt{h} is only possible when having either a coffee \texttt{c} or a publication \texttt{p}.
Propositional logic can be equipped with a Boolean semantics by using $\mathbb{D} = \{0, 1\}$ as domain for all propositions denoting false and true and considering the interpretations $\omega$ as mappings from \symbols to $\semvalues = \{0, 1\}$.
The Boolean semantics $\mu_B$ of propositional logic formulae then follows inductively.
For instance, the above formula $\varphi$ evaluates to $\varphi_B(\omega) = 1 $ for the interpretation
\begin{align}
    \omega(s)
    =
    \begin{cases}
        1 \qquad &\text{if } s \in \left\{\texttt{h}, \texttt{c}\right\}, \\
        0 \qquad &\text{otherwise}.
    \end{cases}
\end{align}

\end{example}

\begin{example}[Fuzzy propositional logic]
Propositional logic can also be equipped with a fuzzy semantics by setting the sets \semvalues and $\mathbb{D}$ to the real unit interval $\left[0, 1\right]$ and using fuzzy operators 
such as the continuous T-norms~\citep{gupta1991theory} to inductively define its semantics $\mu_F$. 

For instance, if one uses \L ukasiewicz fuzzy logic~\citep{lukasiewicz1956investigations} with the T-conorm $\min(1, x + y)$ for the disjunction and the negation $1 - x$,
the formula $\varphi$ (Equation~\ref{eq:main_example}) evaluates to $\varphi_F(\omega) = \min(1, 1 - \omega(\texttt{h}) + \min(1, \omega(\texttt{c}) + \omega(\texttt{p}))) = 1$ for the interpretation
\begin{align}
    \omega(s)
    =
    \begin{cases}
        0.5     \qquad &\text{if } s \in \left\{\texttt{c}, \texttt{p}\right\}, \\
        1       \qquad &\text{otherwise}.
    \end{cases}
\end{align}
That is, having a fuzzy feeling of $0.5$ for having a coffee and having a publication is enough to satisfy $\varphi$ and allow for happiness, while Boolean semantics require absolute certainty on having a coffee or a publication. 
\end{example}


\begin{example}[Linear SMT logic]
The language of linear SMT logic is comprised of sentences over symbols that appear in linear arithmetic operators to form linear arithmetic comparisons that can be connected with logical operators.
For example, one can rewrite the propositional sentence of Equation~\ref{eq:main_example} as the linear SMT formula
\begin{align}
    \texttt{h = 1} \Rightarrow \texttt{(c + p >= 0)},
\end{align}
where \texttt{h}, \texttt{c} and \texttt{p} are symbols with domain $\{-1, 1\}$.
SMT logic can also be given a Boolean semantics by mapping arithmetic comparisons to the set $\semvalues = \{0, 1\}$ by choosing $1$ for interpretations, i.e. assignments of symbols, that satisfy the comparison according to arithmetic and $0$ otherwise.
\end{example}

Numerous other logics exist, such as first-order and temporal logics~\citep{de2013linear}, and their semantics can also be defined in terms of assigning values to symbols or sequences of symbols (for temporal logics).
Similar analyses can be made for automata~\citep{vardi2005automata,manginas2024nesya} and other formal languages~\citep{sun2024neurosymbolic}.

Logical inference can  be viewed as inferring whether there are interpretations that satisfy certain constraints w.r.t. their semantic values.
For instance, SAT-solvers address the question whether there exists an interpretation $\omega$ that satisfies a formula $\varphi$, i.e. that satisfies $\varphi_B(\omega) = 1$.  
In fuzzy logic, one might be interested in considering only those interpretations $\omega$ that satisfy $\varphi_F(\omega) > \tau$.
For these reasons, it will be convenient to introduce logic functions.

\begin{definition}(Logic function)
A \emph{logic function} $l$ is a function that takes a formula $\varphi$ and interpretation $\omega$ and returns a non-zero semantic value only when the value of $\varphi(\omega)$ is contained in a desired subset $\semvalues_l$ of the semantic values \semvalues. 
That is, it is any function $l: L \times \Omega \rightarrow \semvalues$ with $l(\varphi, \omega) = 0$ if $\varphi(\omega) \notin \semvalues_l$.
We call the set $\semvalues_l$ the \emph{selection values} of the logic function.
\end{definition}


The intuition behind a logic function is that it outputs desired semantic values for selected interpretations of interest based on a logical formula. 
For most NeSy AI systems, especially those based on Boolean logic,
$l(\varphi,\omega) = \varphi(\omega)$.
It is only when working with both thresholds and fuzzy logic that more complex logic functions might be necessary, such as $l(\varphi_F,\omega) = \ive{\varphi_F(\omega) > \tau}$, where $\ive{}$ denotes the Iverson brackets that evaluate to 1 if its argument evaluates to true, and yields the value 0 otherwise.

\section{Towards Neurosymbolic Models and Inference}

Neurosymbolic AI models extend logic with neural networks and are, as we will show, typically based on two components: a (possibly fuzzy) logic and a (possibly probabilistic) neural belief~\citep{marra2024statistical}.
While the semantic value $\varphi(\omega)$ of a logical formula in an interpretation is captured by the semantic function $\mu(\varphi,\omega)$, the belief component can be captured by a \emph{belief function} $b_{\params}(\varphi, \omega)$.
The belief $b_\theta(\varphi,\omega)$ can be interpreted as the weight indicating the degree of belief that the interpretation $\omega$ satisfies the formula $\varphi$.
It generally takes the form of a parametrised function
$
b_{\boldsymbol{\theta}}: L \times \Omega
\rightarrow
\mathbb{R}
$
with parameters $\params$ that takes an assignment of symbols and a sentence and outputs the belief.
This belief is often probabilistic or neurally implemented, which means that $\params$ are the parameters of a neural network or a graphical model

\begin{example}[Parametrising a Boolean propositional sentence]
\label{ex:boolean_parametrisation}
Consider again the sentence
$
    \texttt{h} \Rightarrow \texttt{(c} \lor \texttt{p)}
$
from Example~\ref{example:boolean_propositional}.
Now assume there is a neural network that takes as input an image taken from a camera in your local mathematics department.
As output, the network with parameters $\params$ returns three probabilities, one probability $p_{\boldsymbol{\theta}, s}$ for each symbol $s \in \{\texttt{h}, \texttt{c}, \texttt{p}\}$.
Together, assuming \texttt{h}, \texttt{c} and \texttt{p} are independent, these probabilities can be combined into a simple probabilistic belief function $b_{\params}(\varphi, \omega)$ by taking the product of the probabilities, i.e.
\begin{align}
    \label{eq:boolean_parametrisation}
    b_{\boldsymbol{\theta}}(\varphi, \omega)
    =
    \prod_{s \in \{\texttt{h}, \texttt{c}, \texttt{p}\}}
    p_{\boldsymbol{\theta}, s}^{\omega(s)}
    \cdot
    (1 - p_{\boldsymbol{\theta}, s})^{1 - \omega(s)}.
\end{align}
\end{example}

\begin{definition}[Neurosymbolic  model]
\label{def:model}
A neurosymbolic AI model $(L, \mu, \Omega, b_{\params})$ consists of 
a logical language $L$ with a semantics $\mu$ over interpretations $\Omega$ and a belief function $b_{\params}$ with parameters $\params$.
\end{definition}

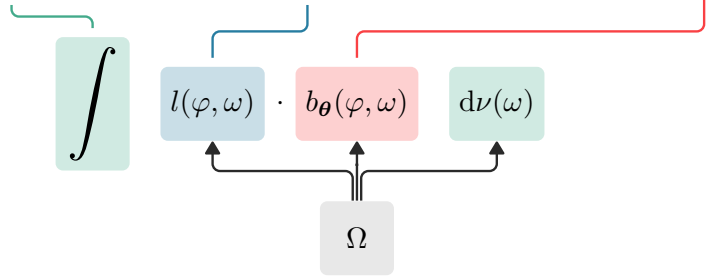
\begin{figure}[t]
    \centering
    \resizebox{\linewidth}{!}{
    \begin{tikzpicture}[remember picture]
\node[sqnode, fill=jungle_green] (integral) {\LARGE$\displaystyle\int$};
\node[sqnode, right=1em of integral.east, fill=celadon_blue] (query) {$l(\varphi, \omega)$};
\node[sqnode, right=1em of query.east, fill=red_salsa] (parametrisation) {$b_{\params}(\varphi, \omega)$};
\node[between=query.east and parametrisation.west] (product) {$\cdot$};
\node[sqnode, right=1em of parametrisation.east, fill=jungle_green] (measure) {$\mathrm{d}\nu(\omega)$};

\node[between=integral and measure] (center) {};
\node[sqnode, below=2em of parametrisation] (interpretations) {$\Omega$};
\node[between=query.south and interpretations.north] (lowercenter) {};

\node[above=2em of query] (text) {
    \textsf{\fontseries{l}\selectfont
        Neurosymbolic inference \colournode{white}{\colouraccent{jungle_green}{\fontseries{m}\selectfont aggregates}}{filter}{0em}\colournode{white}{\colouraccent{celadon_blue}{\fontseries{m}\selectfont logically selected interpretations}}{inconsistent}{0em} from a \colournode{white}{\colouraccent{red_salsa}{\fontseries{m}\selectfont neural belief}}{belief}{0em}.
    }
};

\node[between=filter.south and query.north] (middle) {};
\node[between=filter.south and integral.north] (othermiddle) {};
\end{tikzpicture}
    \begin{tikzpicture}[overlay, remember picture]
        \draw[thinline, draw=celadon_blue] (inconsistent.south) -- (inconsistent.south |- middle) -| (query.north);
        \draw[thinline, draw=red_salsa] (belief.south) -- (belief.south |- middle) -| (parametrisation.north);
        \draw[thinline, draw=jungle_green] (filter.south) -- (filter.south |- othermiddle) -| (integral.north);

        \draw[thinpath] ([xshift=-0.5mm]interpretations.north) -- ([xshift=-0.5mm]interpretations.north |- lowercenter) -| (query.south);
        \draw[thinpath] (interpretations.north) -- (interpretations.north |- lowercenter) -| (parametrisation.south);
        \draw[thinpath] ([xshift=0.5mm]interpretations.north) -- ([xshift=0.5mm]interpretations.north |- lowercenter) -| (measure.south);
    \end{tikzpicture}
    }
    \caption{
        The main intuition behind neurosymbolic inference.
        Note how the interpretations $\Omega$ of the logical language form the interface between neural and symbolic components.
    }
    \label{fig:definition_overview1}
\end{figure}

Neurosymbolic AI models are used to perform inference.
We view inference in a neurosymbolic model $(L, \mu, \Omega, b_{\params})$ as computing the integral graphically illustrated in Figure~\ref{fig:definition_overview1}.
Neurosymbolic inference can be formally defined through neurosymbolic functionals via measure theory and Lebesgue integration (Appendix~\ref{app:formal_functional}).

\begin{definition}[Neurosymbolic  Inference]
\label{def:nesy}
Given a neurosymbolic model $(L, \mu, \Omega, b_\theta)$, a logic function $l$ and a measure space $(\Omega, \Sigma_\Omega, m)$, neurosymbolic inference is defined as computing the result of the following \emph{neurosymbolic functional}
\begin{align}
    \label{eq:nesy_def}
    F_{\boldsymbol{\theta}}(\varphi)
    =
    \int_{\Omega'}
    l(\varphi, \omega)
    \
    b_{\boldsymbol{\theta}}(\varphi, \omega)
    \ \mathrm{d}m(\omega),
\end{align}
where $\Omega'\subseteq \Omega$ is a subset of all possible interpretations determined by a subset of the symbols of $L$.
\end{definition}

The technical intuition for using Lebesgue integration and measures is given in the next paragraph, which can be skipped for the less technically inclined.

Lebesgue integration generalises the usual Riemannian integration to domains that are not real vector spaces, such as the space of interpretations $\Omega$, by using \emph{measures}.
A measure $m$ does nothing else than define a notion of size or volume by taking a set $S$ as input and returning a value $m(S)\in \mathbb{R}$ indicating how large that set is.
Such a notion of volume then forms the basis for integration as integration intuitively aggregates function values over  infinitesimally small pieces of volume.
For instance, the usual Riemann integral $\int_\mathbb{R} f(x)\ \mathrm{d}x$ over the real line $\mathbb{R}$ aggregates function values $f(x)$ over infinitesimally small intervals,
so the length of the interval can be seen as the measure or notion of volume of Riemann integration.
Since the Riemann integral is defined using volumes,
it uses the differential notation $\mathrm{d}x$ to denote infinitesimal pieces of volume or measurements.
In contrast, the Lebesgue integral makes the measure $m$ explicit by using the notation $\mathrm{d}m(x)$ instead (Equation~\ref{eq:nesy_def}).

This definition of neurosymbolic inference generalises weighted model counting (WMC)~\citep{chavira2008probabilistic} and weighted model integration (WMI)~\citep{belle2015probabilistic} beyond purely probabilistic neurosymbolic models~\citep{de2023neural} through the generality of measure theory and Lebesgue integration.
WMC and WMI are recovered from Equation~\ref{eq:nesy_def} by having a finite space of interpretations $\Omega$ with a counting measure (WMC) or an infinite space of interpretations with a measure that combines counting with Riemann integration (WMI).

Importantly, Definition~\ref{def:nesy} leads to precise conditions under which neurosymbolic inference is well-defined.

\begin{proposition}[Well-defined neurosymbolic inference]
Let $(L, \mu, \Omega, b_{\params})$ be a neurosymbolic AI model,  $l$ a logic function and $(\Omega, \Sigma_\Omega, \mathrm{d}m)$ a measure space for which the logic function $l$ and belief function $b_{\boldsymbol{\theta}}$ are measurable,
then neurosymbolic inference is well-defined.
\end{proposition}

The choice of how to parametrise the belief $b_{\params}$ is completely free and does not necessarily have to involve neural networks.
In fact, if one foregoes the use of neural networks and considers probabilistic beliefs, then our definition of a neurosymbolic model reduces to a definition for statistical relational AI (StarAI) \citep{DeRaedtKerstingEtAl16} models.
Consequently, our view on neurosymbolic inference provides a formal framework for inference in StarAI as well.
Such a framework for StarAI was also missing, which shows the utility of a unifying and formal definition.

The question of how to perform learning from the quantities inferred by a neurosymbolic functional can be answered in many different ways.
In settings where the belief is parametrised by neural networks, one can define a loss function in terms of the neurosymbolic functional and learn via backpropagation of this loss.
In other settings, such as in StarAI, different approaches to learning like expectation maximisation can also be used.
In general, our framework does not impose any restrictions on how to perform learning.  We only define the semantics of NeSy models and of inference, not of learning, as is usual when defining semantics.

\section{Neurosymbolic inference unifies neurosymbolic AI}

We will now show that our definitions of neurosymbolic models and inference unifies many prominent neurosymbolic AI frameworks.
These frameworks can be characterised based on their language, semantics and parametrisation (Table~\ref{tab:dimensions_long}).
We will mainly separate systems based on their semantics.

\begin{table}[ht]
    \centering
    \caption{
        How to recover inference in popular neurosymbolic frameworks as neurosymbolic inference following Definition~\ref{def:nesy}.
        (B) indicates the semantics is Boolean in the sense that they are based on true and false values while (F) similarly indicates fuzzy semantics.
        A semantics indicated with (N) is a semantics that uses neural networks to compute more procedural semantics.
        Belief parametrisations can take many forms, but we specifically indicate wether they are deep (D), probabilistic (P) or based on probabilistic circuits (PC)~\citep{choi2020probabilistic}.  
    }
    \resizebox{\linewidth}{!}{
    \begin{tabular}{ccccc}
        \\
        \textbf{System}
            & \textbf{Language}
            & \textbf{Semantics}
            & \textbf{Belief function}
            & \textbf{Logic function}
            \\
        \midrule
        $\alpha$ILP~\citep{shindo2023alpha}
            & Logic programs
            & Stable models (B)
            & D + P
            & Boolean satisfaction
            \\
        $\delta$ILP~\citep{law:noisy}
            & Logic programs
            & Fuzzy
            & Point prediction (D)
            & Fuzzy satisfaction
            \\
        $\Pi$-NeSy~\citep{baaj2025pi}
            & If-then rules
            & Preferential entailment~\citep{dubois1991brief}
            & Possibilistic (D)
            & Preferential satisfaction
            \\
        DeepProbLog~\citep{manhaeve2018deepproblog}
            & Logic programs
            & Well-founded (B)
            & D + P
            & Boolean satisfaction
            \\
        DeepSeaProbLog~\citep{de2023neural}
            & Distributional logic programs
            & Measure semantics~\citep{dos2024declarative} (B)
            & D + P
            & Boolean satisfaction
            \\
        DeepSoftLog~\citep{maene2023soft}
            & Logic programs + embeddings
            & Well-founded (B)
            & D + P
            & Boolean satisfaction
            \\
        DeepStochLog~\citep{winters2022deepstochlog}
            & Definite clause grammars
            & Stochastic (F)
            & D + P
            & Fuzzy satisfaction
            \\
        DL2~\citep{fischer2019dl2}
            & SMT logic
            & Fuzzy semantics (F)
            & Point prediction (D)
            & Fuzzy satisfaction
            \\
        DLM~\citep{marra2019integrating}
            & First-order logic
            & Boolean (B)
            & D + P
            & Boolean satisfaction
            \\
        DRL~\citep{stoian2025beyond}
            & QFLRA
            & Boolean (B)
            & D + P
            & Boolean satisfaction
            \\
        LRNN~\citep{sourek2018lifted}
            & Definite clause logic
            & Fuzzy (F)
            & Embedded formula (D)
            & Fuzzy satisfaction
            \\
        LTN~\citep{serafini_logic_2016}
            & First-order logic
            & Real logic (F)
            & Embedded formula (D)
            & Fuzzy satisfaction
            \\
        NeuPSL~\citep{pryor2022neupsl}
            & Logic programs
            & \L ukasiewicz (F)
            & D + P
            & Fuzzy satisfaction
            \\
        Neural LP~\citep{yang2017differentiable}
            & Polytree DKGs
            & Stochastic (F)
            & D + P
            & Fuzzy satisfaction
            \\
        NeurASP~\citep{yang2020neurasp}
            & Logic programs
            & Stable models (B)
            & D + P
            & Boolean satisfaction
            \\
        NLM~\citep{NLM}
            & First-order rules
            & Neural (N)
            & Embedded formula (D)
            & Function evaluation
            \\
        NLog~\citep{tsamoura2021neural}
            & Logic programs
            & Boolean (B)
            & D + P
            & Boolean satisfaction
            \\
        NLProlog~\citep{weber2019nlprolog}
            & Horn clauses
            & Fuzzy (F)
            & D
            & Fuzzy satisfaction
            \\
        NMLN~\citep{marra2021neural}
            & First-order logic
            & Boolean (B)
            & D + P
            & Boolean satisfaction
            \\
        NTP~\citep{rocktaschel2017end}
            & Function-free first-order rules
            & Neural (N)
            & Embedded formula (D)
            & Function evaluation
            \\
        SBR~\citep{diligenti2017sbr}
            & First-order logic
            & Fuzzy (F)
            & Point prediction (D)
            & Fuzzy satisfaction
            \\
        Scallop~\citep{li2023scallop}
            & Logic programs
            & Provenance (B)
            & D + P
            & Algebraic satisfaction
            \\
        Semantic Loss~\citep{xu2018semantic}
            & Propositional logic
            & Boolean (B)
            & D + P
            & Boolean satisfaction
            \\
        SLASH~\citep{skryagin2022slash}
            & Logic programs
            & Stable models (B)
            & D + P + PC
            & Boolean satisfaction
            \\
        SPL~\citep{ahmed2022semantic}
            & Propositional logic
            & Boolean (B)
            & PC
            & Boolean satisfaction
            \\
        TensorLog~\citep{cohen2020tensorlog}
            & Polytree DKGs
            & Stochastic (F)
            & D + P
            & Fuzzy satisfaction
            \\
    \end{tabular}
    }
    \label{tab:dimensions_long}
\end{table}

\subsection{Neurosymbolic AI with Boolean semantics}

Boolean logic is fundamental to computer science and is also the foundation of a series of neurosymbolic systems with a \emph{probabilistic interpretation}, such as DeepProbLog~\citep{manhaeve2018deepproblog}, Neural Markov Logic Networks (NMLN)~\citep{marra2021neural}, Semantic Probabilistic Layers (SPL)~\citep{ahmed2022semantic} and NeurASP~\citep{yang2020neurasp}.
All of these systems differ in their choice of language, parametrisation or implementation of Boolean semantics (Table~\ref{tab:dimensions_long}), e.g. logic programs with stable model semantics~\citep{gelfond1988stable} (NeurASP) or propositional logic with Boolean semantics (SPL).
However, they are similar in that they are all based on computing probabilities of sentences being true or false.
Returning to our running example, each of the systems can compute the probability of the sentence \q{happiness is only possible when having coffee or a publication} encoded in their respective languages.

\begin{example}[Probabilistic Boolean neurosymbolic AI]
Assuming an independent factorisation $b_{\boldsymbol{\theta}}$ (Equation~\ref{eq:boolean_parametrisation}), the probability of the sentence $\texttt{h} \Rightarrow \texttt{(c} \lor \texttt{p)}$ being true is by definition
\begin{align}
    \int_{\mathbb{B}^3}
        \mu_B(
            \texttt{h} \Rightarrow \texttt{(c} \lor \texttt{p)}
            ,
            \omega
        )
    \
    \cdot
    \prod_{s \in \{\texttt{h}, \texttt{c}, \texttt{p}\}}
    p_{\boldsymbol{\theta}, s}^{\omega(s)}
    \cdot
    (1 - p_{\boldsymbol{\theta}, s})^{1 - \omega(s)}
    \
    \mathrm{d}m_C(\texttt{h}, \texttt{c}, \texttt{p})
\end{align}
where $m_C$ is a counting measure over the 8 possible binary interpretations.
That is, we have neurosymbolic inference with the Boolean semantics $\mu_B$ as logic function and a probabilistic belief function.
As the space of interpretations $\Omega$ is finite, this expression is equivalent to the WMC instance
\begin{align}
    \sum_{\omega \in \mathcal{M}}
    \prod_{s \in \{\texttt{h}, \texttt{c}, \texttt{p}\}}
    p_{\boldsymbol{\theta}, s}^{\omega(s)}
    \cdot
    (1 - p_{\boldsymbol{\theta}, s})^{1 - \omega(s)},
\end{align}
where $
\mathcal{M}
=
\left\{
    \omega \mid 
    \mu_B(
        \texttt{h} \Rightarrow \texttt{(c} \lor \texttt{p)},
        \omega
    )
    =
    1
\right\}
$
is the set of \emph{models} of the sentence $\texttt{h} \Rightarrow \texttt{(c} \lor \texttt{p)}$.
Each of the different Boolean neurosymbolic systems would perform this exact computation, but implemented in their own language and semantics.
NMLN would directly encode the sentence in first-order logic while SPL encodes sentences in propositional logic.
For DeepProbLog and NeurASP, the sentence $\texttt{h} \Rightarrow \texttt{(c} \lor \texttt{p)}$ would be encoded as a logic program with its corresponding well-founded or stable models semantics.
\end{example}

In general, neurosymbolic systems based on Boolean semantics perform inference according to Definition~\ref{def:nesy}.

\begin{claim}
Inference in typical neurosymbolic systems based on Boolean semantics corresponds to neurosymbolic inference of the form
\begin{align}
    \int_{\Omega_B}
    l(\varphi, \omega)
    \
    b_{\boldsymbol{\theta}}(\varphi, \omega)
    \
    \mathrm{d}m_B(\omega),
\end{align}
where $\Omega_B = \mathbb{B}^\atoms \times \mathbb{D}^{\symbols \setminus \atoms}$ and $m_B$ is a combination of binary counting measures and Borel measures.
\end{claim}

\begin{argument}
We show that this statement holds for DeepProbLog, SPL, NeurASP and NMLNs.
The foundational inference task in these systems is computing the probability that a sentence $\varphi$ is true, i.e.
\begin{align}
    \mathbb{P(\varphi)}
    =
    \int_{\Omega_B}
    \varphi_B(\omega)
    b_{\params}(\varphi, \omega)
    \
    \mathrm{d}m_B(\omega).
\end{align}
Hence, the logic function $l$ for DeepProbLog, SPL, NeurASP and NMLNs is equal to the Boolean value $\varphi_B(\omega)$ of the sentence $\varphi$ in the interpretation $\omega$.
The belief function $b_{\params}$ for all systems has to be a probability distribution, yet the form differs per system.
We split up three cases for
\begin{enumerate*}[label=\textbf{(\arabic*)}]
    \item DeepProbLog and NeurASP,\label{item:dplnasp}
    \item SPL, and\label{item:spl}
    \item NMLNs.\label{item:nmln}
\end{enumerate*}

\ref{item:dplnasp} DeepProbLog and NeurASP choose an independently factorising probability distribution as belief.
That is, their belief function is
\begin{align}
    b_{\params}(\varphi, \omega)
    =
    \prod_{s \in \symbols}
    p_{\boldsymbol{\theta}, s}^{\omega(s)}
    \cdot
    (1 - p_{\boldsymbol{\theta}, s})^{1 - \omega(s)},
\end{align}
where $p_{\boldsymbol{\theta}, s}$ is the probability that the binary symbol $s$ is true.~\footnote{Other related systems~\citep{de2023neural} also allow for categorical or continuous distributions resulting in products of probability mass functions or probability densities.}

\ref{item:spl} SPL allows to parametrise the belief as a conditional probabilistic circuit~\citep{shao2020conditional} that is \emph{compatible}~\citep{choi2020probabilistic} with the logical formula $\varphi$.

\ref{item:nmln} NMLNs see the sentence $\varphi$ as a first-order theory $\bigwedge_{i=1}^N \varphi_i$ consisting of $N$ sentences.
Their belief function is then constructed as the normalised exponentiated sum
\begin{align}
    b_{\params}(\varphi, \omega)
    =
    \frac{1}{Z}
    e^{\sum_{i = 1}^N
        \lambda_{\params, i}\cdot \varphi_{i, B}(\omega)
    }
    =
    \frac{1}{Z}
    \prod_{i = 1}^N
    e^{\lambda_{\params, i}\cdot \varphi_{i, B}(\omega)}
    ,
\end{align}
where each $\lambda_{\params, i}$ is a parametrised weight and $Z$ is the normalising constant over interpretations $\Omega$.
\end{argument}

In the probabilistic setting where the belief $b_{\boldsymbol{\theta}}(\varphi, \omega)$ is a probability distribution over the set of interpretations $\Omega$, neurosymbolic inference becomes an instance of either weighted model counting (WMC)~\citep{chavira2008probabilistic} or weighted model integration (WMI)~\citep{belle2015probabilistic} depending on whether $\Omega$ is finite or infinite.

\subsection{Neurosymbolic AI with fuzzy semantics}

Fuzzy semantics for logical languages~\citep{ruspini1991semantics} has enjoyed a lot of interest as a finer-grained alternative to the traditional Boolean semantics.   
While Boolean semantics is two-valued based on absolute truth and falsehood, fuzzy semantics is infinite-valued and expresses a degree of truth by mapping symbols and sentences to the real unit interval.
For neurosymbolic AI, the real-valued nature of fuzzy semantic values can result in a differentiable notion of satisfiability that makes the integration with neural networks easier.
However, it does lead to more diverse computations as different systems can be interested in different restrictions of the fuzzy values of a sentence.

\begin{example}[Fuzzy neurosymbolic AI]\label{ex:fuzzy_nesy}
Many fuzzy neurosymbolic systems only compute the fuzzy value of a sentence given a single fuzzy interpretation.
For example, Logic Tensor Networks (LTN) proposes an intricate way of parametrising the belief $b_{\boldsymbol{\theta}}$ for a single fuzzy interpretation $\omega_{\params}$ by mapping symbols an operators to tensors and tensor operations followed by computing fuzzy values of sentences in the interpretation $\omega_{\params}$.
In case of our running example with the \L ukasiewicz T-norm, the belief $b_{\boldsymbol{\theta}}$ of LTN would be a Dirac delta distribution $\delta$ that gives a single fuzzy value for \texttt{h}, \texttt{c} and \texttt{p} while ignoring all the other possible fuzzy interpretations in the space $\Omega = \left[0, 1\right]^3$.
This construction leads to neurosymbolic inference of the form
\begin{align}
    \int_{\left[0, 1\right]^3} 
    \min(1, 1 - \omega(\texttt{h}) + \min(1, \omega(\texttt{c}) + \omega(\texttt{p})))
    \delta(\omega - \omega_{\boldsymbol{\theta}})
    \
    \mathrm{d}\texttt{h}\mathrm{d}\texttt{c}\mathrm{d}\texttt{p}
    =
    \varphi_F(\omega_{\boldsymbol{\theta}})
    ,
\end{align}
where now $\mathrm{d}\texttt{h}$, $\mathrm{d}\texttt{c}$ and $\mathrm{d}\texttt{p}$ are the usual Borel measure on $\left[0, 1\right]$.
This expression uses fuzzy evaluation as logic function and collapses to the fuzzy value $\varphi_F(\omega_{\boldsymbol{\theta}})$ because of the Dirac delta distribution $\delta$.
\end{example}

In general, our definition of neurosymbolic inference encompasses inference in typical fuzzy neurosymbolic systems.
While covering LTN~\citep{LTN} and SBR~\citep{diligenti2017sbr} inference requires rewriting their inferred quantities, the rewrite exposes connections to other fuzzy systems like NeuPSL~\citep{pryor2022neupsl}.

\begin{claim}
Inference in typical neurosymbolic systems based on fuzzy semantics is neurosymbolic inference of the form
\begin{align}
    \int_{\Omega_F}
    l(\varphi, \omega)
    b_{\boldsymbol{\theta}}(\varphi, \omega)
    \ \mathrm{d}m_F(\omega),
\end{align}
where  $\Omega_F = \left[0, 1\right]^{\atoms}\times\mathbb{D}^{\symbols\setminus \atoms}$ and $m_F$ is a Borel measure.
\end{claim}

\begin{argument}
We prove this claim for the cases of
\begin{enumerate*}[label=\textbf{(\arabic*)}]
    \item logic tensor networks (LTN) and semantic-based regularisation (SBR), and\label{item:ltnsbr}
    \item neural probabilistic soft logic (NeuPSL).\label{item:neupsl}
\end{enumerate*}

\ref{item:ltnsbr} LTN and SBR both compute the fuzzy value $\varphi_F(\omega_{\params})$ of a sentence $\varphi$ in a single parametrised fuzzy interpretation $\omega_{\params}$.
Only considering a single fuzzy interpretation corresponds to choosing a belief function that is a Dirac delta distribution, i.e. $b_{\params}(\varphi, \omega) = \delta(\omega - \omega_{\params})$.
Indeed, we can use the collapsing property of the Dirac delta distribution $\delta$ to write
\begin{align}
    \varphi_F(\omega_{\params})
    =
    \int_{\Omega_F}
    \varphi_F(\omega)
    \delta(\omega - \omega_{\params})
    \
    \mathrm{d}\omega,
\end{align}
where we used notation $\mathrm{d}\omega$ because we can use a traditional Riemann integral.
Hence, LTN and SBR both have the fuzzy value $\varphi_F(\omega)$ as logic function and a Dirac delta as belief function.

\ref{item:neupsl} NeuPSL sets the belief function to be the probability distribution
\begin{align}
    b_{\params}(\varphi, \omega)
    =
    \frac{1}{Z}
    e^{\sum_{i = 1}^N
        \lambda_{\params, i}\cdot \varphi_{i, F}(\omega)
    }
    =
    \frac{1}{Z}
    \prod_{i = 1}^N
    e^{\lambda_{\params, i}\cdot \varphi_{i, F}(\omega)}
    ,
\end{align}
similarly to NMLNs, but with fuzzy semantics.
Its choice of logic function changes from task to task, though NeuPSL generally computes fuzzy expected values.
For instance, the usual expected fuzzy value would use fuzzy satisfaction $\varphi_F(\omega)$ as logic function.
\end{argument}

Note how NeuPSL combines fuzzy logic with a parametrised probabilistic belief over all fuzzy interpretations.

\begin{example}[Probabilistic fuzzy neurosymbolic AI]\label{ex:neupsl}
Systems like NeuPSL relax the hard, Boolean semantic values of atomic expressions to soft, fuzzy values in \L ukasiewicz logic and define a probability distribution $p(\varphi, \omega)$ over the space of fuzzy interpretations for each formula $\varphi$.
This construction allows computing \emph{fuzzy expectations}, e.g. the expectation of the fuzzy value of the sentence $\texttt{h} \Rightarrow \texttt{(c} \lor \texttt{p)}$
\begin{align}
    \Exp{
        \omega \sim p(\varphi, \omega)
    }{
        \varphi_F(\omega)
    }
    =
    \int_{\Omega_F} 
    \min(1, 1 - \omega(\texttt{h}) + \min(1, \omega(\texttt{c}) + \omega(\texttt{p})))
    p(\varphi, \omega)
    \
    \mathrm{d}\omega.
\end{align}
This quantity can then be used to optimise the fuzzy value of a sentence \emph{in expectation} instead of only relying on a point estimate as LTN or SBR does.
\end{example}

Given the example of NeuPSL, it seems fuzzy neurosymbolic systems can become more expressive by parametrising the continuum of fuzzy interpretations.
Indeed, instead of parametrising a probability distribution over all fuzzy interpretations, one can use any other expressive belief function that covers more than one fuzzy interpretation.
For instance, to maintain a completely fuzzy approach, one could turn the set $\Omega_F$ of fuzzy interpretations itself into a fuzzy set by parametrising a membership function $f_m: \Omega_F \rightarrow \left[0, 1\right]$ that corresponds to defining a fuzzy belief function.

\subsection{Limitations}

Our definition of neurosymbolic models (Definition~\ref{def:model}) and inference (Definition~\ref{def:nesy}) make, as we have shown, abstraction of the alphabet soup in neurosymbolic AI.
They have been designed to strike a good balance between generality and mathematical complexity.
While certain edge cases exist that do not directly fit Definition~\ref{def:nesy}, we believe it would not be too hard to extend our definitions to accommodate them.
For instance, maximum a posteriori (MAP) inference, which aims to return the most likely interpretation, would require composing Equation~\ref{eq:nesy_def} with a maximisation operation.
Such edge cases can be covered again by considering nesting or compositions of Equation~\ref{eq:nesy_def}, but would be somewhat more involved.



For similar reasons we have also not given much attention to the choice or construction of the measure in definition~\ref{def:nesy} as the measure is connected to the type of inference.
For example, tasks like MAP can be cast as instances of algebraic model counting (AMC)~\citep{kimmig_algebraic_2017} that aggregate using general algebraic operations, but the measures used in our definitions only consider the usual algebraic structure on the reals $\mathbb{R}$.
However, this problem can be solved by using generalised measures~\citep{wang2010generalized, choquet1954theory, sugeno1972fuzzy}, as in fuzzy measure theory~\citep{zadeh1978fuzzy}.
For ease of exposition, we leave a more detailed analysis using generalised measures for future work.

Finally, we want to draw attention to the fact that our definitions do not limit the semantics of a neurosymbolic model to be strictly logical. 
That is, it is allowed to consider neurosymbolic models with unconstrained or more operational notions of semantics, e.g. NLM~\citep{NLM} and NTP~\citep{rocktaschel2017end} (Table~\ref{tab:dimensions_long}).
This freedom allows us to be inclusive, but opens up the debate on what it means for a system to be really symbolic, which usually requires a logical semantics.


\section{Related work}
This is not the first attempt to arrive at a synthesis and a framework for neurosymbolic AI. 
For instance,~\citet{odense2025semantic} introduce a semantic framework for {\em encoding} logics into neural networks.
However, their emphasis is on the necessary conditions under which a class of neural networks and logical systems can be said to be semantically equivalent.
That is, a neural network can be encoded as a logical theory and the other way around.
This is in line with  Henry Kautz's category {$\mathbf{Neural}_\mathbf{Symbolic}$} to produce a neural network from logical rules.
Our semantics instead focuses on making the logic and belief functions explicit while being rather implicit about the neural network architecture, which is more in line with Henry Kautz's category {$\mathbf{Neural} \mid \mathbf{Symbolic}$} where both the logical and neural components remain and one is not reduced to the other. 

Another related work is ULLER~\citep{van2024uller}, which proposes a unified language for learning and reasoning.
It aims at the \qq{frictionless sharing of knowledge} across neurosymbolic systems and is intended as an interface language, or even interface system, for contemporary neurosymbolic AI systems.
Unlike our approach, ULLER is built upon a fixed first order logic and assigns a concrete semantics for Boolean, fuzzy and probabilistic instances.
In contrast, our framework is not looking for a lingua franca for neurosymbolic AI, but rather focuses on defining a wide variety of neurosymbolic models and inference tasks in a mathematically uniform way.

Other noteworthy approaches include~\citet{van_bekkum_modular_2021}, who show how to combine and visualize specific design patterns of learning and reasoning architectures,~\citet{dash2021tell} who characterise NeSy systems by input formats and loss functions,~\citet{slusarz2023logic} and~\citet{marra2024statistical} who devise various dimensions of neurosymbolic and statistical relational AI systems on which our definition builds.
While these are important developments that can eventually lead to an extensive taxonomy of neurosymbolic AI systems, our definitions identify the essential concepts of neurosymbolic models and show how these concepts define abstract neurosymbolic inference tasks.

\section{Conclusion}

Motivated by the wide range of existing neurosymbolic AI models and approaches, which appear 
quite different on the surface level, we proposed a general and unifying definition of inference in neurosymbolic AI systems
that integrate neural networks with logics.
In our view, neurosymbolic inference consists of computing 
an integral over a product of a logic function and a belief function.  
We provided evidence that our framework is general in that is makes abstraction of prominent contemporary systems such as LTNs, NeuPSL, SBR, SPL, DeepProbLog, NeurASP and NMLNs.

We believe that our definition will be useful for developing both the theory of neurosymbolic AI
by providing a computational framework for designing, evaluating and comparing different neurosymbolic AI systems and tasks,
and for studying their computational and mathematical properties.
We also believe it will be useful for developing an operational framework and system in which many existing neurosymbolic AI systems can be emulated.~\footnote{\url{https://research.kuleuven.be/EU/p/he/p1/erc/deeplog}}~\footnote{\url{https://wms.cs.kuleuven.be/cs/onderzoek/deeplog}}

\section*{Acknowledgements}

This work project has received funding from the European Research Council (ERC) under the European Union’s
Horizon 2020 research and innovation programme (Grant agreement No. 101142702),
the Flemish Government under the “Onderzoeksprogramma Artificiële Intelligentie (AI) Vlaanderen” programme,
the Flemish research foundation (FWO) project “Neurosymbolic AI and Constraint Learning” (Project G047124N)
and the Wallenberg AI, Autonomous Systems and Software Program (WASP) funded by the Knut and Alice Wallenberg Foundation.
The authors would also like to thank the DeepLog team (David Debot, Gabriele Venturato, Giuseppe Marra, Jaron Maene, Lucas Van Praet, Pedro Zuidberg Dos Martires, Rik Adriaensen, Robin Manhaeve, Stefano Colamonaco and Vincent Derkinderen) for the many interesting discussions and feedback on earlier drafts of the paper.

\newpage

\bibliography{ref,MyLibrary,ourbib15}

\newpage

\appendix

\section{Basic of measure theory}
\label{app:formal_functional}

Our definition of neurosymbolic inference uses certain basic concepts from measure theory that we outline here.

\begin{definition}[$\sigma$-algebra and measurable spaces]
Let $X$ be a set, then a \emph{$\sigma$-algebra} $\Sigma$ on $X$ is a non-empty collection of subsets of $X$ that satisfies the properties
\begin{enumerate}
    \item $\forall S \in \Sigma: S^c \in \Sigma$,
    \item $\forall (S_n)_{n\in\mathbb{N}}: (\forall i \in \mathbb{N}: S_i \in \Sigma) \Rightarrow \bigcup_{n \in \mathbb{N}} S_n \in \Sigma$,
    \item $\forall (S_n)_{n\in\mathbb{N}}: (\forall i \in \mathbb{N}: S_i \in \Sigma) \Rightarrow \bigcap_{n \in \mathbb{N}} S_n \in \Sigma$.
\end{enumerate}
That is, a $\sigma$-algebra $\Sigma$ is closed with respect to taking the complement, countable unions and countable intersections.
The elements of $\Sigma$ are called \emph{measurable sets}.
If $\Sigma$ is a \sigmaalg on the set $X$, then the couple $(X, \Sigma)$ is called a \emph{measurable space}.
A function $f$ between two measurable spaces $(S, \Sigma_S)$ and $(T, \Sigma_T)$ is called \emph{measurable} if $f^{-1}(T) \in \Sigma_S$ for each $T \in \Sigma_T$.
\end{definition}

\begin{example}[A \sigmaalg for the Boolean interpretations of propositional logic]
\label{ex:sigma_alg}
Assume we limit the set $A$ of atomic expressions of the language of propositional logic to be finite, e.g. the modern Latin alphabet.
In this case, the set of all possible Boolean interpretations is isomorphic to $\mathbb{B}^{26}$.
Any finite set can easily be provided with a \sigmaalg by taking the powerset of that set, so a \sigmaalg of the set $\mathbb{B}^{26}$ of interpretations could be $\mathcal{P}(\mathbb{B}^{26})$.
It is trivial to verify that this collection indeed satisfies the necessary conditions to be a \sigmaalg of $\mathbb{B}^{26}$.
\end{example}

\begin{definition}[Measure and measure spaces]
Let $(X, \Sigma)$ be a measurable space, then a function $m: \Sigma \rightarrow \mathbb{R} \cup \left\{\infty\right\}$ is called a \emph{measure} if it satisfies
\begin{enumerate}
    \item $m(\varnothing) = 0$,
    \item \textbf{Non-negativity}: $\forall S \in \Sigma: m(S) \geq 0$,
    \item \textbf{Sigma-additivity}:
    $\forall (S_n)_{n \in \mathbb{N}}:
    (\forall i, j, l \in \mathbb{N}: S_i \in \Sigma
    \land
    S_j \cap S_l = \varnothing) \Rightarrow \sigma\left(\bigcup_{n \in \mathbb{N}} S_n\right) = \sum_{n \in \mathbb{N}} m(S_n)$.
\end{enumerate}
In other words, a measure is a positive map of subsets of $X$ to the extended real number line that \q{commutes} with countable unions.
If $m$ is a measure on the measurable space $(X, \Sigma)$, then the triple $(X, \Sigma, m)$ is called a \emph{measure space}.
\end{definition}

\begin{example}[A measure for the Boolean interpretations of propositional logic]
\label{ex:measure}
Assume the same setting as in Example~\ref{ex:sigma_alg} and take the measurable space $(\Omega, \mathcal{P}(\Omega))$ with $\Omega = \mathbb{B}^{26}$.
A well-known measure for finite measurable spaces is the \emph{counting measure} $m_C$ that outputs the cardinality of each element of $\Sigma$, i.e. $m_C(S) = \left|S\right|$.
\end{example}

Given a measure space, one can define a notion of integration that generalises the traditional Riemann integral on real-valued measure spaces to other measure spaces.
This notion of integration is based on the \emph{Lebesgue integral} and is constructed by first considering the family of \emph{simple functions}.

\begin{definition}[Simple function]
If $(X, \Sigma, m)$ is a measure space, then a simple function $s(\boldsymbol{x})$ is a linear combination of indicator functions over disjoint measurable sets with finite measure, i.e.
\begin{align}
    s(\boldsymbol{x})
    =
    \sum_{i = 1}^N
    a_i \cdot \indicator_{S_i}(\boldsymbol{x})
    ,
\end{align}
where $(S_i)_{i = 1}^N$ is a finite sequence of disjoint sets with $S_i \in \Sigma$ and $m(S_i) < \infty$, $a_i \in \mathbb{R}$ and $\indicator_{S_i}$ is the indicator function on the set $S_i$.
That is,
\begin{align}
    \indicator_S(\boldsymbol{x})
    =
    \begin{cases}
        1     \qquad &\text{if } \boldsymbol{x} \in S, \\
        0     \qquad &\text{otherwise}.
    \end{cases}
\end{align}
\end{definition}

The Lebesgue integral is first defined for the family of simple functions, since they inherently grasp the property of linearity that we are used to from the Riemann integral.
Then, the Lebesgue integral for any measurable function $f: X \to \mathbb{R}$ can be defined as the integral of the simple function \q{closest from below} to $f$.

\begin{definition}[Lebesgue integral]
The Lebesgue integral of a simple function
$
s(\boldsymbol{x})
=
\sum_{i = 1}^N
a_i \cdot \indicator_{S_i}(\boldsymbol{x})
$
over a measure space $(X, \Sigma, m)$ is defined as
\begin{align}
    \int
    s
    \ \mathrm{d}m
    =
    \int
    s(\boldsymbol{x})
    \ \mathrm{d}m(\boldsymbol{x})
    =
    \sum_{i = 1}^N
    a_i
    \cdot
    m(S_i).
\end{align}
Moreover, the Lebesgue integral of a \emph{non-negative} measurable function $f: X \to \mathbb{R}$ is then defined as
\begin{align}
    \int
    f
    \ \mathrm{d}m
    =
    \int
    f(\boldsymbol{x})
    \ \mathrm{d}m(\boldsymbol{x})
    =
    \sup\left(
        \{
        \int
        s
        \ \mathrm{d}m
        \mid
        s \leq f
        \text{ and $s$ is simple}
        \}
    \right).
\end{align}
If $f$ is not non-negative, then we define the two non-negative functions $f^+ = \max(0, f)$ and $f^- = \min(0, f)$ such that
\begin{align}
    \int
    f
    \ \mathrm{d}m
    =
    \int
    f^+
    \ \mathrm{d}m
    -
    \int
    f^-
    \ \mathrm{d}m.
\end{align}
\end{definition}

\end{document}